\documentclass{article}
\usepackage{spconf,amsmath,graphicx,booktabs,multirow,booktabs,afterpage}
\usepackage{url}
\usepackage{marvosym}
\usepackage[table]{xcolor}
\title{FadeMem: Biologically-Inspired Forgetting for Efficient Agent Memory}
\name{
Lei Wei\textsuperscript{1,2*},
Xiao Peng\textsuperscript{1}\textsuperscript{\Letter*},
Xu Dong\textsuperscript{2*},
Niantao Xie\textsuperscript{2},
Bin Wang\textsuperscript{2}
}

\address{\textsuperscript{1}Alibaba International Digital Commerce Group \\
\textsuperscript{2}School of Software and Microelectronics, Peking University
}
\begin{document}
\ninept
\maketitle
\begingroup
\renewcommand{\thefootnote}{} % 无编号脚注
\footnotetext{\Letter\ Corresponding author.}
\footnotetext{* These authors contributed equally to this research.}
\endgroup

\begin{abstract}
Large language models deployed as autonomous agents face critical memory limitations, lacking selective forgetting mechanisms that lead to either catastrophic forgetting at context boundaries or information overload within them. While human memory naturally balances retention and forgetting through adaptive decay processes, current AI systems employ binary retention strategies that preserve everything or lose it entirely. We propose FadeMem, a biologically-inspired agent memory architecture that incorporates active forgetting mechanisms mirroring human cognitive efficiency. FadeMem implements differential decay rates across a dual-layer memory hierarchy, where retention is governed by adaptive exponential decay functions modulated by semantic relevance, access frequency, and temporal patterns. Through LLM-guided conflict resolution and intelligent memory fusion, our system consolidates related information while allowing irrelevant details to fade. Experiments on Multi-Session Chat, LoCoMo, and LTI-Bench demonstrate superior multi-hop reasoning and retrieval with 45\% storage reduction, validating the effectiveness of biologically-inspired forgetting in agent memory systems.
\end{abstract}

\afterpage{%
  \begin{figure*}[t]
    \centering
    \includegraphics[width=0.95\textwidth]{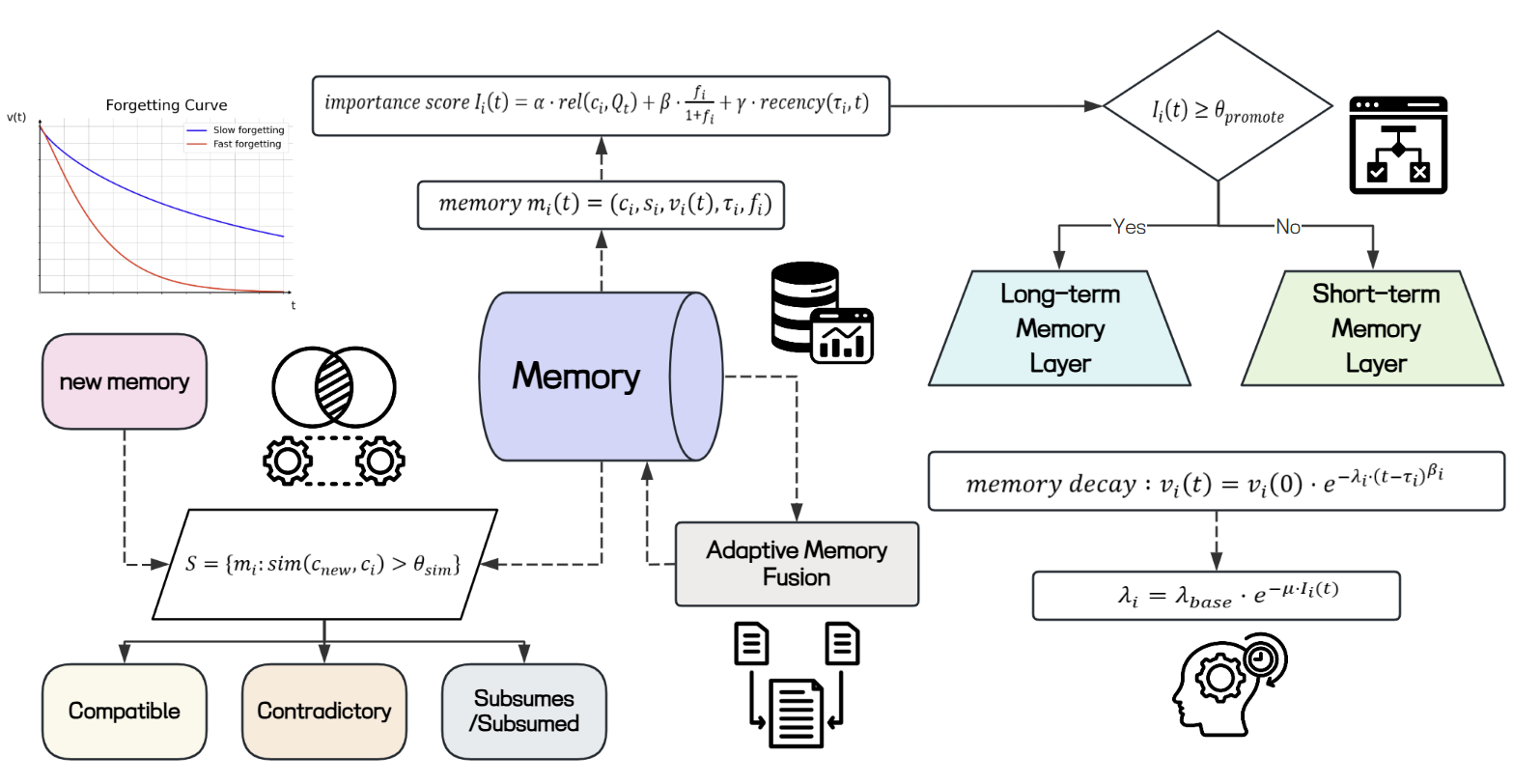}
    \caption{Overview of FadeMem architecture featuring a dual-layer memory hierarchy with adaptive decay mechanisms, LLM-guided conflict resolution, and intelligent memory fusion for efficient long-term agent memory management.}
    \label{fig:Learning to forget}
  \end{figure*}
}

\begin{keywords}
Large Language Models, Agent Memory, Long-Term Context, Adaptive Memory Management
\end{keywords}

\section{Introduction}
The advent of large language models (LLMs) has revolutionized AI systems' ability to process and generate human-like text, yet their practical deployment as autonomous agents remains constrained by fundamental memory limitations \cite{karpukhin2020dense,liu2024lost,kang2026multimodalmultiagentempoweredlegal}. Recent advances in agent memory architectures have explored various approaches to extend context windows and maintain conversation history, from retrieval-augmented generation (RAG) systems that leverage external knowledge bases \cite{lewis2020retrieval,gao2024retrievalaugmentedgenerationlargelanguage} to memory-augmented neural networks that incorporate differentiable memory modules \cite{akbar2025rag,li2021one,lei2025large,martin2022bio}. These developments have enabled agents to handle increasingly complex tasks requiring long-term context retention, multi-turn interactions, and knowledge accumulation across sessions \cite{dai2019transformer,chhikara2025mem0}.

However, existing agent memory architectures suffer from a critical flaw: they lack selective forgetting mechanisms, causing either catastrophic forgetting at context boundaries or information overload within them \cite{kirkpatrick2017overcoming,aleixo2023catastrophic,voina2023biologically,zhang2025survey}. Current agent memory systems predominantly operate on simplistic storage-and-retrieval paradigms that treat all information with equal importance, leading to context windows cluttered with irrelevant details and degraded performance as memory scales. While human memory elegantly balances retention and forgetting through natural decay processes, where unimportant information gradually fades while significant memories are reinforced, current AI systems employ binary retention strategies that either preserve everything within their capacity or lose it entirely \cite{tadros2020biologically,rudroff2024neuroplasticity}. This limitation becomes increasingly problematic as agents handle longer interactions and accumulate vast amounts of potentially redundant or outdated information. The biological inspiration from Ebbinghaus's forgetting curve reveals that human memory strength follows predictable exponential decay patterns modulated by factors such as repetition, emotional significance, and relevance \cite{ebbinghaus1913memory}. This natural forgetting is not a weakness but an adaptive feature that prevents cognitive overload, maintains information relevance, and enables efficient generalization by removing specific details while preserving important patterns \cite{wixted2004psychology}.

To address these fundamental limitations, we propose FadeMem, a biologically-inspired agent memory architecture that incorporates active forgetting mechanisms to mirror human cognitive efficiency \cite{liu2025neural,zheng2025machine}. Our system implements differential decay rates across a dual-layer memory hierarchy, where each memory's retention is governed by adaptive exponential decay functions modulated by semantic relevance, access frequency, and temporal patterns \cite{kang2025memory}. Through LLM-guided conflict resolution and intelligent memory fusion, our architecture naturally consolidates related information while allowing irrelevant details to fade, achieving a dynamic balance between memory capacity and retrieval precision. Our contributions can be summarized as follows: (1) We present the first dual-layer biologically inspired agent memory with adaptive forgetting.
(2) We devise a unified framework with LLM-guided conflict resolution and memory fusion that enforces temporal consistency and aggressively compresses redundancy, yielding compact, coherent memory states.
(3) Extensive experiments on Multi-Session Chat, LoCoMo, and LTI-Bench show superior multi-hop reasoning and retrieval with large storage savings, and thorough ablations validate each component’s impact.

\section{Methodology} 
\subsection{Dual-Layer Memory Architecture with Differential Forgetting} Inspired by human memory systems, we design a dual-layer architecture that mimics the differential forgetting rates observed in biological memory, as shown in Fig.~\ref{fig:Learning to forget}. Each memory $m_i$ at time $t$ is represented as: \begin{equation} m_i(t) = (c_i, s_i, v_i(t), \tau_i, f_i) \end{equation} where $c_i$ is the content embedding, $s_i$ is the original text, $v_i(t) \in [0, 1]$ is the memory strength, $\tau_i$ is the creation timestamp, and $f_i$ is the access frequency. The memory importance score determines layer assignment: \begin{equation} I_i(t) = \alpha \cdot \text{rel}(c_i, Q_t) + \beta \cdot \frac{f_i}{1 + f_i} + \gamma \cdot \text{recency}(\tau_i, t) \end{equation} where $Q_t$ represents recent context, the frequency term follows a saturating function to prevent over-weighting, and recency is defined as $\text{recency}(\tau_i, t) = \exp(-\delta(t - \tau_i))$. In practice, we replace the raw count $f_i$ with an exponentially time-decayed access rate $\tilde f_i=\sum_j \exp(-\kappa (t-t_j))$ to emphasize recent accesses. Memories are dynamically assigned to layers based on importance: \begin{itemize} \item Long-term Memory Layer (LML): High-importance memories with slow decay \item Short-term Memory Layer (SML): Low-importance memories with rapid decay \end{itemize} Layer transitions occur when: \begin{equation} \text{Layer}(m_i) = \begin{cases} \text{LML} & \text{if } I_i(t) \geq \theta_{\text{promote}} \\ \text{SML} & \text{if } I_i(t) < \theta_{\text{demote}} \end{cases} \end{equation} This allows memories to migrate between layers as their importance evolves over time. The thresholds $\theta_{\text{promote}}$ and $\theta_{\text{demote}}$ are determined through grid search on validation data. Using $\theta_{\text{promote}}>\theta_{\text{demote}}$ introduces hysteresis that prevents oscillation. \subsection{Biologically-Inspired Forgetting Curves} We measure time in days, consistent with our 30-day evaluation setup. We model memory decay using differential exponential functions that simulate human forgetting patterns, consistent with Ebbinghaus's forgetting curve: \begin{equation} v_i(t) = v_i(0) \cdot \exp\left(-\lambda_i \cdot (t - \tau_i)^{\beta_i}\right) \end{equation} The decay rate adapts to memory importance: \begin{equation} \lambda_i = \lambda_{\text{base}} \cdot \exp(-\mu \cdot I_i(t)) \end{equation} where $\lambda_{\text{base}}$ approximates human short-term memory decay rates and $\mu$ modulates the importance effect. The shape parameter $\beta_i$ depends on the memory layer: \begin{equation} \beta_i = \begin{cases} 0.8 & \text{if } m_i \in \text{LML} \quad \text{(sub-linear decay)} \\ 1.2 & \text{if } m_i \in \text{SML} \quad \text{(super-linear decay)} \end{cases} \end{equation} These values mirror biological memory consolidation where long-term memories exhibit slower, more gradual decay. Memory consolidation occurs during access, simulating the strengthening effect observed in human memory: \begin{equation} v_i(t^+) = v_i(t) + \Delta v \cdot (1 - v_i(t)) \cdot \exp(-n_i/N) \end{equation} where $\Delta v$ is the base reinforcement strength, $n_i$ counts accesses within a sliding window of $W$ days, and $N$ implements diminishing returns consistent with spacing effects in human learning. Memories undergo automatic pruning when their strength falls below $\epsilon_{\text{prune}}$ or they remain dormant beyond $T_{\text{max}}$ days. 

\noindent\textbf{Half-life}~~With time measured in days, the half-life of memory $m_i$ under our model is \[ t_{1/2}(i)=\Big(\tfrac{\ln 2}{\lambda_i}\Big)^{\!1/\beta_i},\quad \lambda_i=\lambda_{\text{base}}\exp(-\mu I_i(t)). \] At $I_i(t)=0$, this gives $t_{1/2}\approx 11.25$ days for LML ($\beta_i=0.8$) and $t_{1/2}\approx 5.02$ days for SML ($\beta_i=1.2$) when $\lambda_{\text{base}}=0.1$. \subsection{Memory Conflict Resolution} When new information arrives, we detect and resolve conflicts through semantic analysis and LLM-based reasoning. For each new memory $m_{\text{new}}$, we retrieve semantically similar memories: \begin{equation} \mathcal{S} = \{m_i : \text{sim}(c_{\text{new}}, c_i) > \theta_{\text{sim}}\} \end{equation} where $\text{sim}(\cdot)$ is cosine similarity on $\ell_2$-normalized embeddings. For each $m_i \in \mathcal{S}$, an LLM examines $(s_{\text{new}}, s_i)$ and classifies their relationship into one of four categories in text: \emph{compatible}, \emph{contradictory}, \emph{subsumes}, or \emph{subsumed}. We then apply the corresponding resolution strategy: 

\textbf{Compatible}: Both memories coexist, while the existing memory’s importance is reduced by redundancy: \begin{equation} I_i = I_i \cdot \bigl(1 - \omega \cdot \text{sim}(c_{\text{new}}, c_i)\bigr). \end{equation} 

\textbf{Contradictory}: Apply competitive dynamics where newer information suppresses older. We use a window-normalized age difference with $W_{\text{age}}$ days: \begin{equation} v_i(t) = v_i(t)\cdot \exp\!\Bigl(-\rho \cdot \mathrm{clip}\!\bigl((\tau_{\text{new}}-\tau_i)/W_{\text{age}},\,0,\,1\bigr)\Bigr). \end{equation} 

\textbf{Subsumes/Subsumed}: The more general memory absorbs the specific one via LLM-guided merging (content is consolidated; redundant details are compressed). The parameters $\omega$ and $\rho$ control redundancy penalty and suppression strength respectively, calibrated through ablation studies. \subsection{Adaptive Memory Fusion} To maintain efficiency while preserving information integrity, we implement LLM-guided fusion for related memories. Fusion candidates are identified through temporal-semantic clustering: \begin{equation} \mathcal{C}_k = \{m_i : \text{sim}(c_i, c_k) > \theta_{\text{fusion}} \land |\tau_i - \tau_k| < T_{\text{window}}\} \end{equation} where $\theta_{\text{fusion}}$ ensures semantic coherence and $T_{\text{window}}$ maintains temporal locality. For clusters exceeding a size threshold, we perform intelligent fusion via LLM that preserves unique information, temporal progression, and causal relationships. The fused memory inherits aggregated properties: \begin{equation} v_{\text{fused}}(0) = \max_{i \in \mathcal{C}_k} v_i(t) + \epsilon \cdot \text{var}(\{v_i\}) \end{equation} where the strength combines the maximum individual strength with a variance-based bonus, reflecting that diverse supporting memories create stronger consolidation. We clip $v_{\text{fused}}$ to $[0,1]$. \begin{equation} \lambda_{\text{fused}} = \frac{\lambda_{\text{base}}}{1 + \log(|\mathcal{C}_k|)} \end{equation} During fusion we set $\lambda_i \leftarrow \lambda_{\text{base}}\cdot \xi_{\text{fused}}\cdot \exp(-\mu I_i)$ with $\xi_{\text{fused}}=1/(1+\log|\mathcal{C}_k|)$. The reduced decay rate for fused memories reflects their consolidated importance. Information preservation is validated through LLM verification with threshold $\theta_{\text{preserve}}$. If preservation falls below threshold, fusion is rejected. The complete memory evolution follows: \begin{equation} \mathcal{M}_{t+\Delta t} = \text{Fusion}(\text{Resolution}(\text{Decay}(\mathcal{M}_t, \Delta t) \cup \{m_{\text{new}}\})) \end{equation} This creates an adaptive system that naturally forgets unimportant information while strengthening and consolidating important memories, mirroring human cognitive processes. All hyperparameters were determined through systematic ablation studies, balancing retention quality against computational efficiency.

\section{Experiments}

\subsection{Experimental Setup}

\noindent\textbf{Datasets}~~We evaluate our approach on three diverse datasets that capture different aspects of long-term agent memory requirements. We use Multi-Session Chat (MSC) \cite{xu2022beyond} containing 5,000 multi-session dialogues spanning up to 5 sessions per user, with an average context length of 1,614 tokens per session. For long-context evaluation, we employ LoCoMo \cite{maharana2024evaluating}, focusing on multi-hop reasoning across extended contexts. Additionally, we construct a synthetic long-term interaction dataset (LTI-Bench) simulating 30-day agent-user interactions with controlled information evolution patterns, containing 10,780 interaction sequences with explicit temporal dependencies and contradiction scenarios. 

\noindent\textbf{Baselines}~~We compare against three categories of memory management approaches: (1) \textit{Fixed-window methods}: including context windows of 4K, 8K, and 16K tokens with FIFO eviction \cite{dai2019transformer,liu2024lost}; (2) \textit{RAG-based systems}: LangChain Memory with default configurations; (3) \textit{Specialized agent memory}: Mem0 \cite{chhikara2025mem0} as our primary baseline, representing state-of-the-art unified memory layers, and MemGPT \cite{packer2023memgpt} with hierarchical memory management.

\noindent\textbf{Evaluation Metrics.}~~We assess performance across multiple dimensions. For memory efficiency, we measure Storage Reduction Rate (SRR) as $\text{SRR} = 1 - |\mathcal{M}_{\text{retained}}|/|\mathcal{M}_{\text{total}}|$, quantifying the proportion of memory saved through intelligent forgetting. Retrieval quality is evaluated through Relevance Precision@K (RP@K), measuring the precision of top-K retrieved memories. We also compute Temporal Consistency Score (TCS) to assess chronological coherence in memory retrieval, ranging from 0 to 1 where higher values indicate better temporal ordering. For task performance, we report F1 scores on downstream tasks, particularly focusing on multi-hop reasoning capabilities. Additionally, we measure Factual Consistency Rate (FCR) via LLM-based fact checking \cite{manakul2023selfcheckgpt} to ensure memory coherence after updates and conflict resolution. For conflict resolution evaluation, we report accuracy (Acc.) as the rate of correct strategy selection and consistency (Cons.) as the factual coherence maintained post-resolution.

\noindent\textbf{Implementation Details}~~Our system employs GPT-4o-mini \cite{achiam2023gpt} for conflict resolution and memory fusion operations, with embeddings generated using text-embedding-3-small. Key hyperparameters determined through grid search on validation sets include: $\lambda_{\text{base}} = 0.1$, $\theta_{\text{promote}} = 0.7$, $\theta_{\text{demote}} = 0.3$, and $\theta_{\text{fusion}} = 0.75$. All experiments use a dual-layer architecture with maximum capacities of 1,000 memories in LML and 500 in SML. Statistical significance is assessed using paired t-tests with $p < 0.05$.

\subsection{Memory Retention and Forgetting Dynamics}

We evaluate whether our biologically-inspired forgetting effectively balances retention with efficiency using 30-day simulated interactions on LTI-Bench. Table~\ref{tab:retention} shows retention rates for different information categories.

\begin{table}[h]
\centering
\caption{Memory retention analysis on \textbf{LTI-Bench} after 30 days of continuous interaction. Critical facts include user preferences/constraints; contextual info includes topics/states.}
\label{tab:retention}
\begin{tabular}{lccc}
\toprule
Method & Critical Facts & Contextual Info & Storage Used \\
\midrule
Fixed-16K & 50.2\% & 44.8\% & 100\% \\
LangChain & 71.2\% & 65.3\% & 100\% \\
Mem0 & 78.4\% & 69.1\% & 100\% \\
MemGPT & 75.6\% & 62.8\% & 85.3\% \\
\textbf{FadeMem} & \textbf{82.1\%} & \textbf{71.0\%} & \textbf{55.0\%} \\
\bottomrule
\end{tabular}
\end{table}

Our approach achieves 82.1\% retention of critical facts using only 55.0\% storage. Important memories exhibit $3$--$5\times$ slower decay than baseline, with 23\% of low-importance memories promoted to LML based on access patterns.

\subsection{Conflict Resolution Performance}

We inject 4075 controlled conflicts on LTI-Bench across three types to evaluate LLM-guided resolution. Table~\ref{tab:conflict} reports accuracy and post-resolution consistency.

\begin{table}[h]
\centering
\caption{Conflict resolution on \textbf{LTI-Bench} across types. Acc.: correct strategy selection; Cons.: factual coherence after resolution.}
\label{tab:conflict}
\setlength{\tabcolsep}{3.8pt} % tighten column spacing
\small % or \footnotesize
\begin{tabular}{@{}lcccccc@{}}
\toprule
\multirow{2}{*}{Method} & \multicolumn{2}{c}{Contradiction} & \multicolumn{2}{c}{Update} & \multicolumn{2}{c}{Overlap} \\
\cmidrule(lr){2-3}\cmidrule(lr){4-5}\cmidrule(lr){6-7}
 & Acc. & Cons. & Acc. & Cons. & Acc. & Cons. \\
\midrule
FIFO   & -   & 45.2\% & - & 73.1\% & -   & 61.3\% \\
Mem0   & 62.3\% & 71.4\% & 84.7\% & 82.3\% & 45.6\% & 73.8\% \\
MemGPT & 58.7\% & 68.9\% & 81.2\% & 79.6\% & 51.3\% & 75.4\% \\
\textbf{FadeMem}  & \textbf{66.2\%} & \textbf{78.0\%} & \textbf{87.1\%} & \textbf{86.5\%} & \textbf{53.4\%} & \textbf{76.8\%} \\
\bottomrule
\end{tabular}
\end{table}

Our LLM-guided mechanism achieves 68.9\% macro-averaged accuracy and 80.4\% macro-averaged consistency across the three conflict types. The temporal suppression effectively favors recent information while maintaining historical context. 

\subsection{Cross-Dataset Evaluation}

To demonstrate generalizability, we evaluate on MSC for conversational memory and LoCoMo for long-context reasoning. Table~\ref{tab:cross_datasets} shows performance across different evaluation metrics.

\begin{table}[h]
\centering
\caption{Results on \textbf{MSC} and \textbf{LoCoMo}. MSC reports RP@10 and TCS; LoCoMo reports multi-hop F1, Factual Consistency Rate (FCR), and Storage Reduction Rate (SRR).}
\label{tab:cross_datasets}
\begin{tabular}{lccccc}
\toprule
\multirow{2}{*}{Method} & \multicolumn{2}{c}{MSC} & \multicolumn{3}{c}{LoCoMo} \\
\cmidrule(lr){2-3}\cmidrule(lr){4-6}
 & RP@10$\uparrow$ & TCS$\uparrow$ & F1$\uparrow$ & FCR$\uparrow$ & SRR$\uparrow$ \\
\midrule
Fixed-16K & 58.7\% & 0.71 & 5.17 & 78.9\% & 0.00 \\
LangChain & 71.5\% & 0.77 & 25.75 & 81.2\% & 0.00 \\
Mem0      & 74.8\% & 0.79 & 28.37 & 83.6\% & 0.00 \\
MemGPT    & 73.1\% & 0.78 & 9.46  & 82.9\% & 0.15 \\
\textbf{FadeMem} & \textbf{77.2\%} & \textbf{0.82} & \textbf{29.43} & \textbf{85.9\%} & \textbf{0.45} \\
\bottomrule
\end{tabular}
\end{table}

Our approach consistently outperforms baselines on both datasets. On MSC, we achieve 77.2\% RP@10, demonstrating superior retrieval of relevant conversational history. The TCS of 0.82 indicates better temporal coherence across multi-session interactions. On LoCoMo, our multi-hop F1 score of 29.43 shows effective long-context reasoning capabilities, surpassing Mem0 (28.37) and significantly outperforming MemGPT (9.46). Notably, we achieve 85.9\% factual consistency rate while maintaining 45\% storage reduction (SRR=0.45) through intelligent forgetting.

\begin{figure}[h]
\centering
\includegraphics[width=0.48\textwidth]{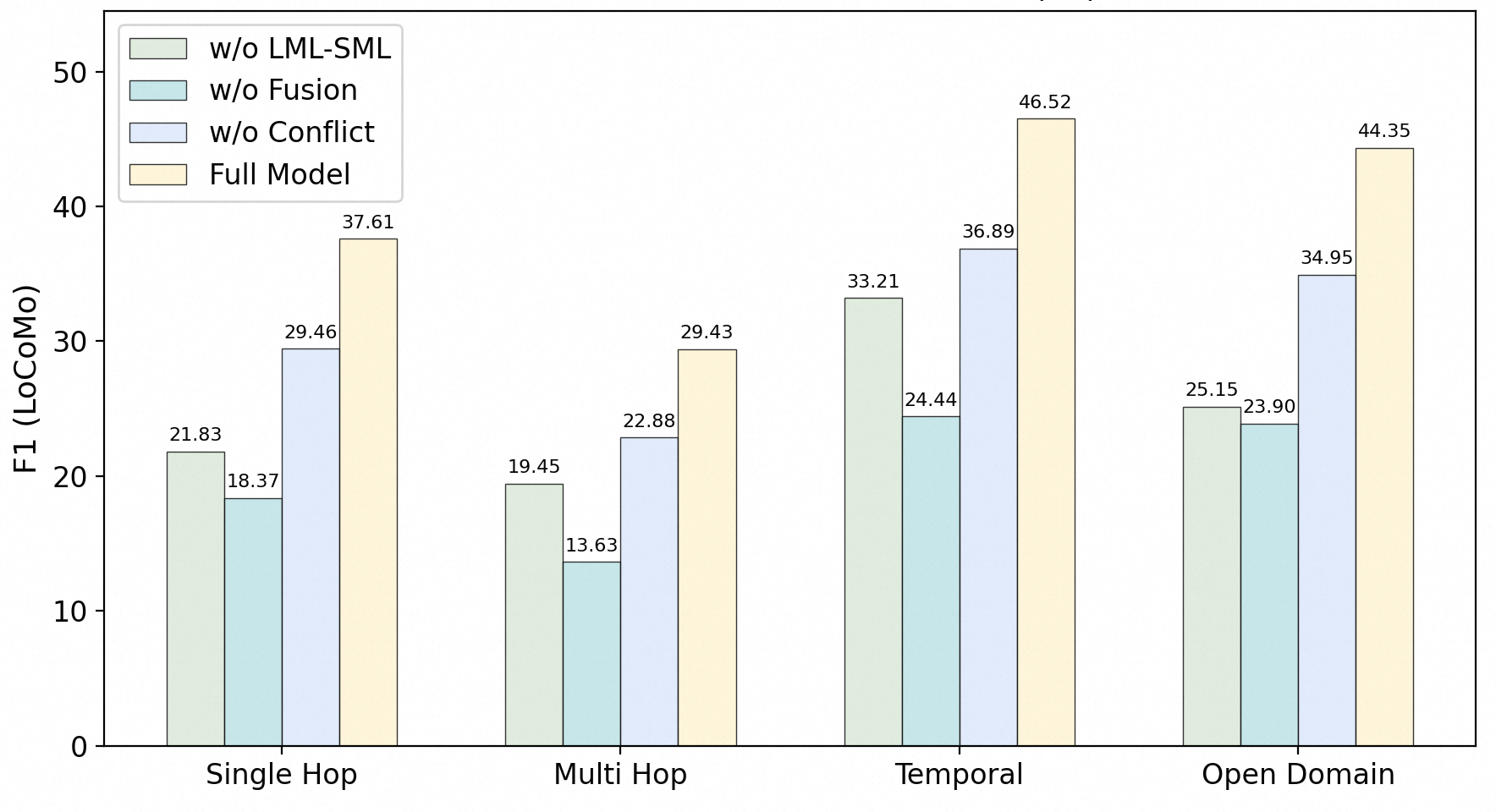}
\caption{Ablation study results on \textbf{LoCoMo} across different task types. Each bar shows F1 scores when removing specific components compared to the full model.}
\label{fig:ablation}
\end{figure}

\subsection{Ablation Study}

We conduct ablation studies on LoCoMo to analyze the contribution of each component in our memory management framework, as illustrated in Fig.~\ref{fig:ablation}.

The ablation results demonstrate the critical importance of each component. Removing the dual-layer architecture (w/o LML-SML) causes significant performance degradation across all tasks, with multi-hop F1 dropping from 29.43 to 19.45 (33.9\% decrease), highlighting its role in effectively separating long-term and short-term memories. The memory fusion component proves essential for maintaining performance, as its removal (w/o Fusion) results in the most severe degradation, with multi-hop F1 plummeting to 13.63 (53.7\% decrease). Conflict resolution (w/o Conflict) also plays a vital role, particularly for maintaining factual consistency—its absence reduces multi-hop F1 to 22.88 (22.4\% decrease). The full model achieves the best performance across all task types, with particularly strong results on temporal tasks (F1=46.52) and open-domain tasks (F1=44.35), demonstrating the synergistic effect of combining all components.

\section{Conclusion}
We presented FadeMem, a biologically-inspired agent memory architecture that introduces adaptive forgetting to LLM-based systems. By implementing differential decay rates modulated by semantic relevance, access frequency, and temporal patterns, combined with LLM-guided conflict resolution and intelligent memory fusion, our approach achieves superior retention of critical information while reducing storage by 45\%. The dual-layer memory hierarchy naturally consolidates related information while allowing irrelevant details to fade, achieving a dynamic balance between memory capacity and retrieval precision. Experiments on Multi-Session Chat, LoCoMo, and LTI-Bench demonstrate consistent improvements in multi-hop reasoning and retrieval performance compared to existing baselines, validating the effectiveness of incorporating human-like forgetting patterns inspired by Ebbinghaus's forgetting curve. This work establishes that selective forgetting, rather than being a limitation, is essential for preventing information overload and maintaining relevance in agent memory systems. Future work will explore meta-learning approaches to further enhance adaptive decay parameters and extend the framework to multi-agent collaborative memory systems.

% Below is an example of how to insert images. Delete the ``\vspace'' line,
% uncomment the preceding line ``\centerline...'' and replace ``imageX.ps''
% with a suitable PostScript file name.
% -------------------------------------------------------------------------

\bibliographystyle{IEEEbib}
\bibliography{strings,refs}

\end{document}